\def\year{2019}\relax
\documentclass[letterpaper]{article} 
\usepackage{aaai19}  
\usepackage{times}  
\usepackage{helvet}  
\usepackage{courier}  
\usepackage{url}  
\usepackage{graphicx}  

\usepackage{booktabs} 
\usepackage{acronym}
\usepackage{booktabs} 
\usepackage{amsfonts}
\usepackage{amssymb}
\usepackage{physics}
\usepackage{braket}
\usepackage{subfigure}
\usepackage{acronym}
\usepackage{amsmath}
\usepackage{url}

\acrodef{ML}[ML]{Machine Learning}
\acrodef{QT}[QT]{Quantum Theory}
\acrodef{PCA}[PCA]{Principal Component Analysis}
\acrodef{IR}[IR]{Information Retrieval}
\acrodef{ANN}[ANN]{Artificial Neural Networks}
\acrodef{QM}[QM]{Quantum Mechanics}
\acrodef{BCIQT}[BCIQT]{Binary Classifier Inspired by Quantum Theory}
\acrodef{SDT}[SDT]{Signal Detection Theory}
\acrodef{NB}[NB]{Na\"{\i}ve Bayes}
\acrodef{SVM}[SVM]{Support Vector Machine}
\acrodef{KNN}[k-NN]{$k$ Nearest Neighbours}
\acrodef{DT}[DT]{Decision Tree}

\begin{document}
%
\title{Binary Classifier Inspired by Quantum Theory}
\author{Prayag Tiwari\\
	Department of Information Engineering\\
    University of Padova\\
	Via Gradenigo 6/b 35131 - Padova, Italy\\
    https://sites.google.com/view/prayag-tiwari/home\\
	tiwari@dei.unipd.it
	\And
    Massimo Melucci\\
    Department of Information Engineering\\
    University of Padova\\
	Via Gradenigo 6/b 35131 - Padova, Italy\\
	http://www.dei.unipd.it/~melo/\\
    melo@dei.unipd.it
}

\maketitle

\begin{abstract}
\ac{ML} helps us to recognize patterns from raw data. \ac{ML} is used in numerous domains i.e. biomedical, agricultural, food technology, etc. Despite recent technological advancements, there is still room for substantial improvement in
  prediction. Current \ac{ML} models are based on classical theories of
  probability and statistics, which can now be replaced by \ac{QT} with the aim of improving the effectiveness of \ac{ML}. In this paper, we propose the \ac{BCIQT} model, which outperforms the state of the art classification in terms of recall for every category. 
\end{abstract}

\noindent

\subsection{Introduction}
\ac{ML} is a set of models which can automatically identify the hidden patterns
in the data and can then utilize hidden patterns to make decisions in condition
of uncertainty.  \ac{ML} has been progressively implemented in several areas
including chemistry, biomedical science and robotics. \ac{ML} falls into three
categories, i.e. supervised learning (e.g. classification), unsupervised
learning (e.g. clustering) and reinforcement learning. In this paper we focus on
classification, which is the way to represent and allocate objects into different
categories.

\ac{QT} is the probabilistic approach to representing and predicting properties
of microscopic phenomena. Given an observable and an arbitrary state
of a microscopic particle, \ac{QT} computes a probability distribution of the
values of the observable. The quantum formalism is explicitly acceptable to
explain distinct types of stochastic processes. Several nonstandard
implementations of the quantum formalism has emerged. For instance, the quantum
formalism have been utilized vastly in the economic processes, game theory  and cognitive
science  as well.

Since the data is growing exponentially, current state-of-the-art models are
still not effective.  In particular, \emph{recall} is still unsatisfactory
because most classification models aim to maximize precision especially when the
items of a class can be ranked by a certain measure of membership to the class;
a glaring example is the search of the Internet.  In contrast, recall is crucial
in many daily tasks aiming to find all the pertinent items of a class such as patent search and biomedical image classification.

Our approach is to develop a new theoretical approach inspired by \ac{QM} in
order to dig into the quantum world and come up with new and effective models
which are capable of increasing recall.  Our hypothesis is that, since \ac{QM} has
already shown its effectiveness in several fields, it may also be effective in
\ac{ML}.  To this end we will exploit Quantum Probability theory, which is the quantum
generalization of classical probability theory and was developed by Von
Neumann. While classical probability theory provides that a system can be in
either state 0 or 1, quantum probability comes into existence to go beyond
classical theory and describes states which can be anything in-between 0 and 1. In
this paper, we propose the \ac{BCIQT} model which is a step towards shifting
from classical models to quantum models.
\section{Proposed Methodology}

\subsection{Classical and Quantum Signal Detection Theory}
\label{sec:quant-detect-fram}

\ac{BCIQT} is based on the overlap between \ac{SDT} and \ac{QM}. The main
difference between the classical framework and the quantum framework of signal
detection regards what encoders encode and what decoders decode
\cite{helstrom1969quantum}.

In the classical framework,
there is c-c (classical-classical) mapping from a symbol to the wave to the
corrupted channel; then the decoder produce c-c (classical-classical) mapping
from the corrupted channel wave to a symbol.  In the quantum framework, there is a coder between the source and
the channel; the classical symbol is transmitted through the quantum
state. Initial encoding starts like c-q (classical-quantum) mapping from the
symbol to the quantum state selected from a finite set of possible states. More
details about classical and quantum \ac{SDT} can be found in
\cite{helstrom1969quantum}.

\subsection{Binary Classifier Inspired by Quantum Theory}

A novel \ac{BCIQT} that is inspired by quantum detection theory is described in
this section.  For each category we supposed that each training sample was about
the category or not.  For a given category and the set of training samples, we
used the projector $\Delta$ for each category to identify whether the test
sample was about the category or not. To determine whether the test sample was
about the category, $\Delta$ was examined against a vectorial representation of
the test samples.

Consider a set of distinct features calculated from the whole sample collection.
Each sample could be represented as a vector of features; each element in the
feature vector was a non-negative number such as frequency.  Each sample in the
training set had a binary label in $\{0,1\}$.  The main goal of \ac{BCIQT} was
to obtain one binary label for each sample in the test set.

The \ac{BCIQT} estimated two density operators $\rho_{0}$ and $\rho_{1}$, one
operator for each category or class and its complement, by using the training
samples; in particular, for each class, the negative training samples were
utilized to estimate $\rho_{0}$ and the positive training samples were utilized
to estimate $\rho_{1}$.

In order to achieve these density operators $\rho_{0}$ and $\rho_{1}$, we first
calculated the total number of samples with non-zero values for each particular
feature.  In such a way, one vector $\ket{v}$ was obtained for each class. Since
we were considering the binary case, two vectors $\ket{v_{0}}$ and $\ket{v_{1}}$
were obtained; the former referred to the negative training samples and the
latter referred to the positive training samples; these vectors may be
considered as statistics of the features in a class.  We normalized the vectors
to obtain $|\braket{v|v}|^2=1$.  Then, we calculated the outer product in order
to obtain the density operators $\rho_{0}$ and $\rho_{1}$ as follows:
\begin{equation}
  \label{eq:rhos}
  \rho_0 = \frac{\ket{v_0} {\bra{v_0}} }{tr(\ket{v_0} {\bra{v_0}})}
  \qquad
  \rho_1 = \frac{\ket{v_1} {\bra{v_1}} }{tr(\ket{v_1} {\bra{v_1}})}
\end{equation}
We computed the projection operator $\Delta$ according to
\cite{melucci2016relevance}, that is,
\begin{equation}
  \label{eq:decomposition}
  \rho_1 - \lambda \rho_0 = \eta\,\Delta + \beta\,\Delta^\perp \qquad \eta > 0
  \qquad \beta < 0 \qquad \Delta\,\Delta^\perp = 0
\end{equation}
where $\xi$ is the prior probability of the negative class and
$ \lambda = {\xi}\,/\,(1-\xi) $; moreover, $\eta$ is the positive eigenvalue
corresponding to $\Delta$ which represents the subspaces of the vectors
representing the sample to be accepted in the target class.

We set $\lambda = 1$ to simply mean that both classes had the same prior
probability ($\xi=0.5$); moreover, there was no cost for wrong detection
$C_{00}=C_{11}=0$; finally, the costs of false alarm and miss were constant
($C_{01}=C_{10}$). Eventually, we determined the binary label for the given test
sample $S_{j}$ by inspecting the value of
$ \langle w_{S_j} \vert \Delta \vert w_{S_j} \rangle$: If
$\langle w_{S_j} \vert \Delta \vert w_{S_j}\rangle \geq 0.5$, then $C(S_j) = 1$;
otherwise $C(S_j) = 0$.

\section{Experiment}

The MNIST database \footnote{\url{http://yann.lecun.com/exdb/mnist/}} of
handwritten digits has a training set of 60,000 examples, and a test set of
10,000 examples. There are 9 categories from 0 to 9 but excluded 9. It is a subset of a larger set
available from the National Institute of Standards and Technology (NIST). The
digits have been size-normalized and centered in a fixed-size image.

The four models i.e. \ac{NB}, \ac{SVM}, \ac{KNN}  and \ac{DT}  were used as baselines. Prior to training the models,
the top 100 features were selected as the best features for all the models in terms of recall. The chi-square feature selection model was used.

We used one-vs-all strategy: for each category, the training samples labeled as
pertinent to the category are considered positive examples, while the rest are
considered negative examples. While training the model, five fold cross
validation was used. As it can be seen from Table \ref{tab:1}, our proposed
model performs better than any state-of-art-model in terms of recall for every
category. By changing number of features, evaluation measures(i.e. accuracy, precision, recall and f-measure) also change and provide comparable results to the baselines.

\begin{table}[h!]
\large
\caption{Comparison of Recall among  k-nearest neighbors(KNN), Decision Tree(DT), Naive Bayes (NB), Support Vector Machine (SVM) and Binary Classifier Inspired by Quantum Theory (BCIQT)}

\begin{tabular}{cccccc}
	\hline
    Category	&	KNN	& DT	&	NB	&	SVM	&  BCIQT	\\
	\hline
    		0	&0.959	&0.884	&0.889	&0.292&	\textbf{1}

	\\
    		1	& 0.699&	0.710&	0.582&	0.390&	\textbf{0.996}

	\\
    		2	&0.704&	0.652&	0.709&	0.474&	\textbf{1}

	\\
    		3	&0.623&	0.508&	0.792&	0.346&	\textbf{0.997}

	\\
    		4	&0.643&	0.621&	0.666&	0.259&	\textbf{0.999}

	\\
    		5	&0.892&	0.855&	0.872&	0.621&	\textbf{1}

	\\
    		6	&0.743&	0.755&	0.873&	0.454&	\textbf{0.999}

	\\
    		7	&0.753&	0.728&	0.779&	0.332&	\textbf{1}

	\\
    		8	&0.749&	0.677&	0.817&	0.209&	\textbf{1}

	\\
	\hline
\end{tabular}
\label{tab:1}
\end{table}

\section{Conclusion and Future Works}

We found out that our proposed model outperforms the state-of-the-art models in terms of
\emph{recall}; therefore, this model can be safely implemented if someone is
looking for high recall. We believe that this is an encouraging result and
opens a gateway towards quantum inspired \ac{ML} approaches. As for  future
work, we would like to develop multi-class classifiers (i.e. how to assign an
item to more than one class) and multi-label classifiers (i.e. how to deal with
non-binary labels), and re-rank the test items of a class by increasing precision as well.

\section{ Acknowledgments}
"This project has received funding from the European Union's Horizon 2020 research and innovation programme under the Marie Sklodowska-Curie grant agreement No 721321".

\bigskip
\noindent 
\bibliography{quantum-theory}
\bibliographystyle{aaai}
\end{document}